\documentclass[sn-mathphys,Numbered,iicol]{sn-jnl}
\usepackage{graphicx}%
\usepackage{multirow}%
\usepackage{amsmath,amsfonts}
\usepackage[title]{appendix}%
\usepackage{xcolor}%
\usepackage{textcomp}%
\usepackage{manyfoot}%
\usepackage{booktabs}%
\usepackage{algorithm}%
\usepackage{listings}%
\usepackage{algorithmic}%
\usepackage{booktabs}%
\usepackage{stfloats}
\usepackage{subfig}
\usepackage{graphicx}
\usepackage{url}
\usepackage{caption}
\usepackage{array}
\usepackage{ragged2e}
\usepackage{verbatim}
\usepackage{graphicx}
\DeclareMathOperator*{\argmax}{arg\,max}

\begin{document}

\title[Learning to bag with a simulation-free reinforcement learning framework for robots]{Learning to bag with a simulation-free reinforcement learning framework for robots}

\author[1]{\fnm{Francisco} \sur{Munguia-Galeano}}\email{MunguiaGaleanoF@cardiff.ac.uk}

\author[2]{\fnm{Jihong} \sur{Zhu}}\email{jihong.zhu@york.ac.uk}

\author[3]{\fnm{Juan David} \sur{Hernández}}\email{HernandezVegaJ@cardiff.ac.uk}

\author*[1]{\fnm{Ze} \sur{Ji}}\email{jiz1@cardiff.ac.uk}

\affil*[1]{\orgdiv{School of Engineering}, \orgname{Cardiff University}, \orgaddress{\city{Cardiff}, \postcode{CF24 3AA}, \country{United Kingdom}}}

\affil[2]{\orgdiv{School of Physics Engineering and Technology}, \orgname{University of York}, \orgaddress{\city{York}, \postcode{YO10 5EZ}, \country{United Kingdom}}}

\affil[3]{\orgdiv{School of Computer Science and Informatics}, \orgname{Cardiff University}, \orgaddress{\city{Cardiff}, \postcode{CF24 4AG}, \country{United Kingdom}}}


\abstract{Bagging is an essential skill that humans perform in their daily activities. However, deformable objects, such as bags, are complex for robots to manipulate. This paper presents an efficient learning-based framework that enables robots to learn bagging. The novelty of this framework is its ability to perform bagging without relying on simulations. The learning process is accomplished through a reinforcement learning algorithm introduced in this work, designed to find the best grasping points of the bag based on a set of compact state representations. The framework utilizes a set of primitive actions and represents the task in five states. In our experiments, the framework reaches a 60\% and 80\% of success rate after around three hours of training in the real world when starting the bagging task from folded and unfolded, respectively. Finally, we test the trained model with two more bags of different sizes to evaluate its generalizability.}





\maketitle

\section{Introduction}

Robots with human-level dexterity that can handle deformable objects may encourage a smoother integration of robots in daily and industrial activities~\cite{culleton2017framework}. In practice, daily activities depend on more than manipulating rigid objects. In this context, the robots' capacity to manipulate deformable objects to operate in human environments is a necessity~\cite{zhu2022challenges}.
Among deformable objects, bags are used in several relevant tasks, such as transporting objects, packing, and shopping. Even though there have been studies on how to manipulate deformable objects, such as paper~\cite{elbrechter2011bi, elbrechter2012folding, balkcom2004introducing, balkcom2008robotic}, fabrics~\cite{borras2020grasping,jangir2020dynamic,hoque2022visuospatial}, ropes~\cite{nair2017combining,sundaresan2020learning,shi2022reactive}, cables~\cite{zhu2019robotic,zhou2020practical} and  meat~\cite{jorgensen2019adaptive}, the problem of learning to bag with robots is still under-explored. 

\begin{figure*}[t]
\centering
\includegraphics[width=5.2in,scale=0.5]{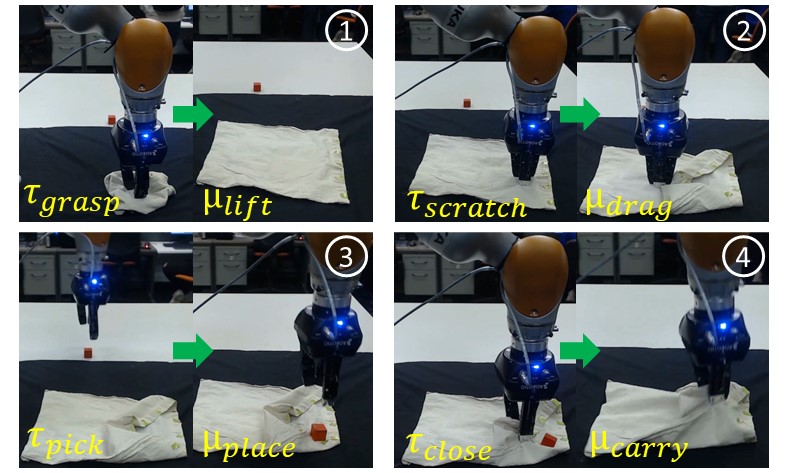}
\caption{The robot, in four steps, performs the bagging task. In the first step, the robot unfolds the bag. In the second step, the bag is opened by the robot. The robot places the red cube in the bag's opening in the third step. In the fourth step, the robot carries the bag, completing the task. }
\label{fig:env}
\end{figure*}

Bagging is a complex task for robots because there exist challenges related to perception, occlusions, modeling of the bag's dynamics, ambiguity related to finding the opening, and how to grasp one or two layers of the bag depending on the current state of the task. Reinforcement learning (RL) has the potential to deal with the problems mentioned above. However, RL agents are commonly trained in simulation, which brings more challenges when implementing the agent in real-world tasks~\cite{sharma2022learning}. One challenge is that before simulating the bag, the model must be the most similar as possible to the real object~\cite{yin2021modeling,polydoros2017survey}. On the other hand, when switching from simulation to real-world, the agent must deal with occlusions, incomplete information, and noises. These are vital factors that make difficult the generalization of RL~\cite{nguyen2019review}.

\begin{figure*}[!t]
\centering
\smallskip
\begin{tabular}{ c c c }
        \begin{tabular}{c}
        \smallskip
        \subfloat[]{\includegraphics[width=1.4in]{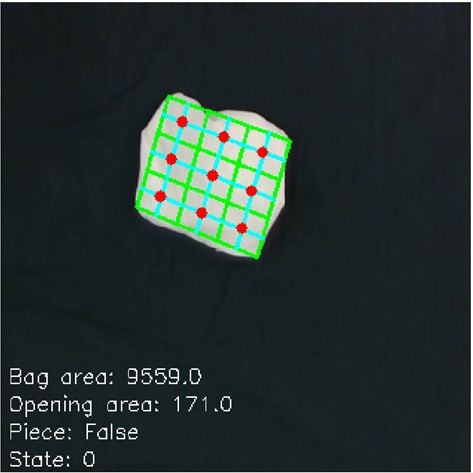}%
        \label{fig:s_a}}\\
        \hfill
        \smallskip
        \subfloat[]{\includegraphics[width=1.4in]{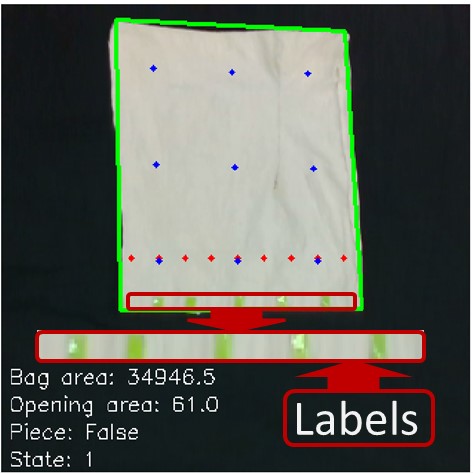}%
        \label{fig:s_b}}\\
        \end{tabular}
        &
        \begin{tabular}{c}
        \smallskip
        \subfloat[]{\includegraphics[width=1.4in]{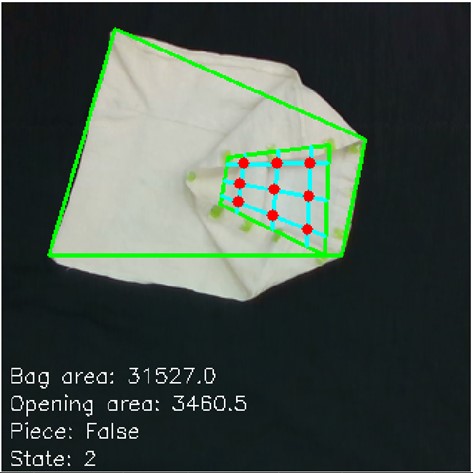}%
        \label{fig:s_c}}\\
        \hfill
        \smallskip
        \subfloat[]{\includegraphics[width=1.4in]{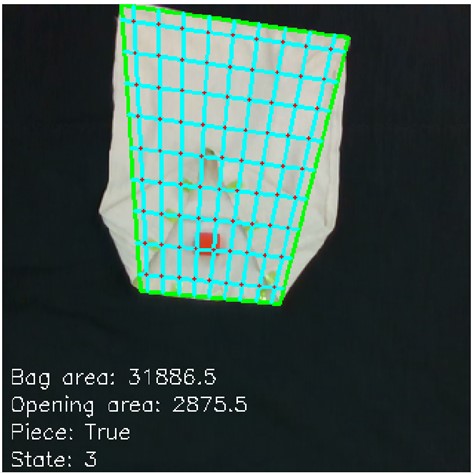}%
        \label{fig:s_d}}\\
        \end{tabular}
        &
        \begin{tabular}{c}
        \smallskip
        \subfloat[]{\includegraphics[width=1.4in]{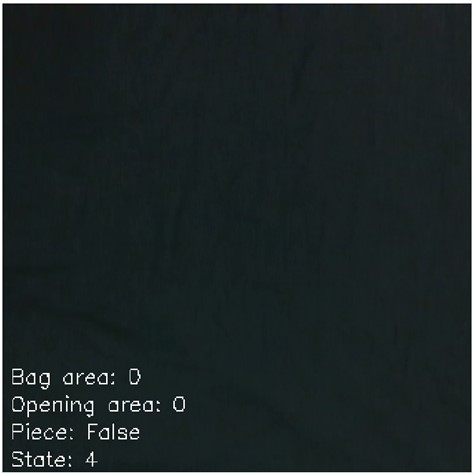}%
        \label{fig:s_e}}\\
        \hfill
        \smallskip
        \subfloat[]{\includegraphics[width=1.4in]{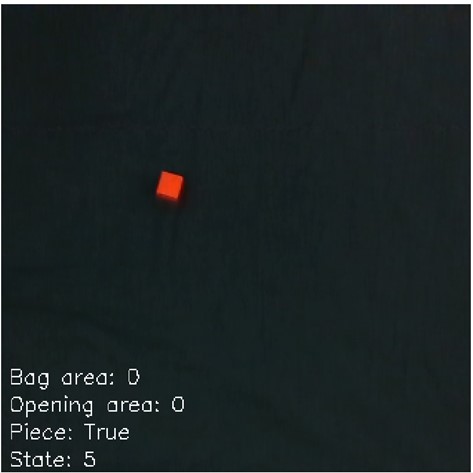}%
        \label{fig:s_f}}\\
        \end{tabular}
\end{tabular}
\caption{This figure illustrates the five states that comprise the bagging task, where the red and blue dots represent the grasping points the robot can select. In (a), the bag is folded such that its area is small, and the opening is not visible. In (b), the bag is unfolded, and the opening is visible. In (c), the bag's opening area is large enough to put an object inside. In (d), the object is in the bag's opening, distinguishing this state from the others. In (e), the task succeeded because no visible objects were left on the table, meaning the robot carried both the bag and the object. Lastly, (f) shows a failure case when the robot took the bag, but the red cube was still on the table.}
\label{fig:perception}
\end{figure*}

This paper presents a learning framework for robot-bagging tasks with compact state representations and primitive actions, aiming to efficiently train a robot to learn bagging in the real world\footnote{Demo available at \tt\url{https://youtu.be/UV9HcEummx0} }. The framework identifies five possible states and utilizes eight primitive actions related to several grasping points on the bag. The task is solved in four steps (Fig. \ref{fig:env}): unfolding, opening, placing the piece, and carrying. The main contributions of this paper are (i) an RL algorithm is introduced, allowing robots to efficiently learn how to bag in the real world, (ii) a versatile state representation for the bagging task, and (iii) a framework that provides a reliable perception of the bag state for learning.

The problem domain to empirically validate the framework is encompassed of a cotton bag and a red cube (Fig. \ref{fig:env}). The framework first learns to perform the task through a different number of steps, as explained in Section~\ref{sec:expsetup}. Then, the trained model is used to perform the task using two more bags with different sizes and positions to test the framework's generalization capabilities.

The rest of this article is structured as follows. First, Section~\ref{sec:rel} summarizes related works in RL and manipulation of deformable objects. The problem formulation is described in Section~\ref{sec:problem}. Then, in Section~\ref{sec:algorithms} formally introduces the framework, followed by Section~\ref{sec:expsetup}, which describes the experimental setup. Section~\ref{sec:results} presents the results, while Section~\ref{sec:discussion} discusses them. Finally, Section~\ref{sec:conclusions} concludes this paper.

\section{Related work}
\label{sec:rel}

One way to tackle the challenge of handling deformable objects with robots is by using simulated environments that are valuable resources for training agents to learn how to solve a given task. For example, Seita et al.~\cite{seita2021learning} used transporter networks to learn how to rearrange deformable objects, where 3D deformable structures such as bags are included. Bahety et al.~\cite{bahety2022bag} proposed a method to rearrange, which is based on two policies learned in simulation. The first policy rearranges the objects, and the second one learns to lift them. A disadvantage of this approach is that it assumes that the bag is always open. Therefore, in a simulated environment, all the information regarding the opening of the bag and the objects is available, which is useful when training an agent requires a large number of episodes to learn the task. However, when running the agent in the real world, the differences between the simulation and the real-world tasks may lead to undesirable and dangerous behaviors. 

In the literature, methods exist to transfer the knowledge obtained during simulation to the real world (simulation-to-reality). Ma et al.~\cite{ma2022learning} utilize a method that sets several grasping points on a cloth surface. A graph neural network uses these points to learn their dynamics. Subsequently, when the task is transferred to the real world, it is easier to track the points than the whole cloth. Xioman et al.~\cite{wang2022learning} introduced a method in which an agent is trained to learn how to wrap boxes in a simulated environment. The actual texture of the deformable object is taken from the real object such that the transition from simulation to the real world is smoother than when taken from the pure simulation. 

In the context of bagging, Iterative Interactive Modeling for Knotting Plastic Bags~\cite{gao2023iterative} is an approach that focuses on the bag's handles, learns from demonstrations, and uses a set of primitive actions to knot the handles. A disadvantage of this approach is that the detection of the handles relies on a large dataset. The approach proposed by Chen et al.~\cite{chen2022autobag} is of particular relevance to this paper. Their algorithm, AutoBag, is built upon a set of primitive actions that involve reorienting plastic bags until the opening becomes visible, allowing objects to be placed inside. The authors' research primarily focuses on plastic bags. Plastic bags tend to keep their shape after manipulation, where the mechanical characteristics are utilized in favor of the manipulation task. For example, when the opening is visible, the bag tends to maintain its shape. In contrast, with a textile-based bag, once the layer is no longer gripped, it falls, necessitating modifications to the AutoBag algorithm to accommodate this behavior. In this context, this work aims to investigate a bagging process involving textile-based bags and a learning-based solution performed in the real world.

Despite the vast literature exploring several RL approaches in simulation, less attention has been paid to RL in the real world due to several problems~\cite{dulac2021challenges}, such as costly robot time, motion constraints, or stochastic behaviors of objects surrounding the robot. Although training RL in simulation is effective, transferring the trained policies to the real world is a challenge that must be taken seriously, often resulting in significant performance degradation due to the simulation-to-reality gap~\cite{zhao2020sim}. Therefore, tackling complex problems such as bagging in the real world may open the door to addressing further challenges in robotics and RL.

\begin{figure*}
\centering
\includegraphics[width=5.2in]{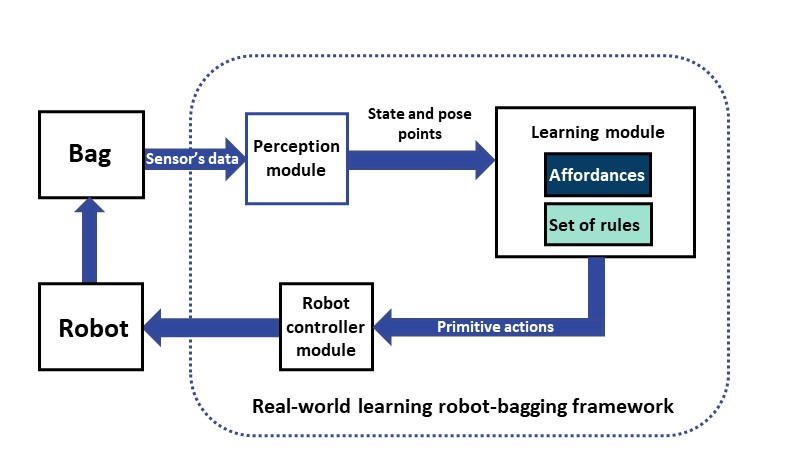}
\caption{Proposed framework for learning to bag using RL. }
\label{fig:framework}
\end{figure*}

\section{Problem formulation}
\label{sec:problem}

We formulated the bagging task as a Markov decision process (MDP)~\cite{sutton1998introduction} and aimed to find a solution using a learning policy $\pi$. In an MDP, the agent executes a valid action $a$ from the set of actions $A$ in the current state $s$ and transitions to a valid state $s'$, where $s$ and $s'$ belong to the set of states $S$, according to the unknown dynamics of the bag. The environment provides a reward according to the reward function $R(s,a)$ upon transitioning to a new state. The set of states $S$ is given by the following:

\begin{equation}\label{eq:1}
\begin{split}
S=\{s_0,s_1,s_2,s_3,s_4,s_5\},
\end{split}
\end{equation}

\noindent where $s_0$ represents the folded bag, $s_1$ is the representation of the bag expanded, $s_2$ represents the bag opened, $s_3$ shows when the red object is on the bag, $s_4$ is the success state, and $s_5$ is the fail state.  More specifically, the states are denoted as follows: 

\begin{itemize}
    \item For $s_0$: Folded bag.

    \textbf{Condition}: the bag's area is small, and the opening is not visible (Fig.~\ref{fig:s_a}).

    \item For $s_1$: Expanded bag.

    \textbf{Condition}: The bag is unfolded, the opening is visible, and the green labels can be seen (Fig.~\ref{fig:s_b}).

    \item For $s_2$: Opened bag.

    \textbf{Condition}: The opening area of the bag is large enough to accommodate an object (Fig.~\ref{fig:s_c}).

    \item For $s_3$: Red object on the bag.

    \textbf{Condition}: The object is in the opening of the bag, which distinguishes this state from the others (Fig.~\ref{fig:s_d}).

    \item For $s_4$: Success state.

    \textbf{Condition}: No visible objects were left on the table, indicating that the robot carried both the bag and the object (Fig.~\ref{fig:s_e}).

    \item For $s_5$: Fail state.

    \textbf{Condition}: The robot took the bag, but the red cube was still on the table (Fig.~\ref{fig:s_f}).

\end{itemize}

Hence, the current state can be calculated as follows:
\begin{scriptsize}
\begin{equation}\label{eq:a1}
\begin{split}
    s= 
\begin{cases}
    s_0,& (A_{\text{bag}} < A_{\text{th}})\wedge (A_{o}=0) \wedge (A_{\text{cube}}=0)\\
    s_1,& (A_{\text{bag}} > A_{\text{th}})\wedge (A_{o}>0) \wedge (A_{\text{cube}}=0)\\
    s_2,& (A_{\text{bag}} > A_{\text{th}})\wedge (A_{o}>A_{\text{oth}})\wedge (A_{\text{cube}}=0)\\
    s_3,& (A_{\text{bag}} > A_{\text{th}})\wedge (A_{o}>A_{\text{oth}})\wedge (A_{\text{cube}}>0)\\
    s_4,& (A_{\text{bag}} = 0) \wedge (A_{o}=0) \wedge (A_{\text{cube}}=0)\\
    s_5,& (A_{\text{bag}} = 0) \wedge (A_{o}=0) \wedge (A_{\text{cube}}>0),
\end{cases}
\end{split}
\end{equation}
\end{scriptsize}

\noindent where $A_{bag}$ is the bag's area, $A_{th}$ is a threshold value indicating a small area, $A_{o}$ is the area of the opening, $A_{oth}$ is the threshold value indicating a large enough opening and $A_{cube}$ stands for the area of the cube. The robot can select among a set of primitive actions executed in pairs. Let:

\begin{equation}\label{eq:2}
\begin{split}
\Phi=\{\langle\tau_{grasp},\mu_{lift}\rangle,\langle\tau_{scract},\mu_{drag}\rangle,\\
\langle\tau_{pick},\mu_{place}\rangle,\langle\tau_{close},\mu_{carry}\rangle \},
\end{split}
\end{equation}

\noindent be the set of tuples that contains the primitive actions $\tau$ that represents the primary primitive action executed by the robot and $\mu$ the complementary primitive actions (Fig.~\ref{fig:setup}). The primitive action $\tau_{grasp}$ is the robot action to grasp two layers of the bag, and $\mu_{lift}$ is when the bag is raised above the table and drops the bag aiming to unfold it. $\tau_{scratch}$ is the action to grasp only one layer of the bag, and the primitive action $\mu_{drag}$ takes place when the robot moves a grasped layer point to a placing point. The $\tau_{pick}$ and $\mu_{place}$ actions refer to picking and placing the red object. $\tau_{close}$ grasps one layer of the bag when the object is placed in the opening, and $\mu_{carry}$ lifts the bag to grab the object. This paper aims to find an optimal policy $\pi^*$ that computes the correct sequence of primitive actions to transition through the states $S$ till solving the bagging task.

\section{Real-world learning robot-bagging framework}
\label{sec:algorithms}

This section presents a framework that aims to solve the problem of learning to bag in the real world with RL (Fig.~\ref{fig:framework}). The framework comprises three modules: perception, learning and robot controller.

\subsection{Perception}

The perception module plays a crucial role in extracting relevant information from the bag, including its area, opening area, grasping points, and current state. It receives data from an Intel® RealSense™ camera, which provides RGB and depth images. To process this data, the module utilizes the OpenCV library. By filtering colors from the black background and obtaining the bag's contours, the module can calculate the $A_{bag}$ value. 

\begin{algorithm}[htbp]
\caption{Opening area's calculator}\label{alg:alg2}
\begin{algorithmic}[1]
\REQUIRE \textit{points}: A list of 2D points
\ENSURE $A_o$: The area of the opening

\IF{$\textsc{CountPoints}(\textit{points}) < 3$}
    \STATE $A_o \gets 0$
    \RETURN $A_o$
\ENDIF

\STATE $\textit{triangles} \gets []$
\STATE $\textit{nodes} \gets \textsc{CountPoints}(\textit{points})$
\STATE $\textit{center} \gets \textsc{GetCenter}(\textit{points})$
\WHILE{\textbf{True}}
    \STATE $\textit{pts} \gets \textit{points}[:]$
    \STATE $i, n \gets \textsc{FindClosestNode}(\textit{center}, \textit{pts})$
    \STATE $j, m \gets \textsc{FindClosestNode}(\textit{n}, \textit{pts})$
    \STATE $\textit{triangles.append}([\textit{center}, \textit{n}, \textit{m}])$
    \STATE $\textit{pts.pop}(j)$
    \STATE $k, o \gets \textsc{FindClosestNode}(\textit{n}, \textit{pts})$
    \STATE $\textit{triangles.append}([\textit{center}, \textit{n}, \textit{o}])$
    \STATE $\textit{points.pop}(i)$
    \IF{$\textsc{CountPoints}(\textit{points}) < 3$}
        \STATE \textbf{break}
    \ENDIF
\ENDWHILE

\STATE $A_o \gets 0$
\FOR{\textbf{each} $\textit{triangle}$ \textbf{in} $\textit{triangles}$}
    \STATE $A_o=A_o+\textsc{GetArea}$
    $(\textit{triangle})$
\ENDFOR

\RETURN $A_o$
\end{algorithmic}
\end{algorithm}

In order to assist with the automatic classification of bag states, green labels have been added around the bag's opening (refer to Fig.~\ref{fig:s_b}). It is worth mentioning that perception is not the main focus of this work. Hence we simplify the experiment configurations in order to reliably obtain the states from observations. Specifically, green markers were used to facilitate the detection of the bag opening. Algorithm~\ref{alg:alg2} is utilized to calculate the opening's area. This is achieved by providing an array of points containing the pair of coordinates obtained from the labels. Then through the generation of triangles, it is possible to sum all their areas and, in this way, obtain the opening's area $A_o$.

This information enables the module to determine whether the bag is on the table, folded, or unfolded. To differentiate the bag from the background, we use a black canvas as the background to make it easier to extract the bag, which is white. Similarly, the object for bagging is a cube in red. In brief, the primary functions of the perception module are:

\begin{itemize}
\item to determine the current state of the task;
\item to provide pose points depending on the state;
\item to measure the bag's current area; and
\item to measure the current opening's area.
\end{itemize}

The perception module generates multiple pose points denoted by $g=\zeta^2$, where $g$ represents the number of pose points and $\zeta$ is the griding parameter. For instance, Fig.~\ref{fig:perception} illustrates a configuration with $g=9$ and $\zeta=3$. Increasing the value of $g$ results in more points available for grasping and lifting the bag, consequently leading to a larger set of actions to explore. Thus, the position of the grasping points can be determined.
 
The $s_0$ state (Fig.~\ref{fig:s_a}) can be distinguished from the others because the opening is not visible, and the bag area is small compared to when it is unfolded. In this state, there are $g=9$ grasping points (red dots in Fig.~\ref{fig:s_a}) whose position can be obtained with the RealSense camera. Those points are stored in the set $P_{s_0}$. Before unfolding the bag and reaching the next state, the robot must explore which one of these grasping points is the best to grasp the bag, lifting it at a given height stored in the $D_{s_0}$ set and dropping it.

In the $s_1$ state (Fig.~\ref{fig:s_b}), the opening is visible, and the bag's area has reached a feasible value for opening the bag. Besides, since the opening area is small, $s_1$ can be distinguished from the other states based on the criteria of the opening area. In this state, the perception module provides $g=9$ grasping points close to the opening (red dots in Fig.~\ref{fig:s_b}) which are stored in the set $P_{s_1}$, and $g$ placing points (blue dots in Fig.~\ref{fig:s_b}) stored in the set $D_{s_1}$. Consequently, the robot's goal in this state is to find which combination of actions over the grasping point maximizes the opening area.

In the $s_2$ state (Fig.~\ref{fig:s_c}), the opening's area serves as a trigger to identify this state. First, the perception module stores the object's position to be bagged in $P_{s_2}$. Then, the perception module sets $g=9$ placing points (red dots in Fig.~\ref{fig:s_c}) and stores them in the set $D_{s_2}$. The robot can place the object to be bagged on one of those placing points and get a reward depending on the closeness with the center of the bag's opening.

In the $s_3$ state (Fig.~\ref{fig:s_d}), the red object is placed in the opening. In this state, the robot is required to explore more grasping points. For this reason, the number of pose points is $g=81$ (red dots in Fig.~\ref{fig:s_d}). Then, the pose points are stored in the set $P_{s_4}$ while the lifting poses corresponding to the complementary primitive action in this state are stored in the set $D_{s_4}$. Therefore, the robot can interact with the bag and decipher which grasping point allows it to finish the task. When the robot lifts the bag, and the module can not detect any object left, as shown in Fig.~\ref{fig:s_d}, state 4 $s_4$ is identified, and the task is finished. On the other hand, when there are still objects on the table, the module identifies state 5 $s_5$, which is a failed attempt to bag the object (Fig.~\ref{fig:s_f}). 

\subsection{Learning}

The learning module seeks to find the optimal combination of grasping points on the bag and primitive actions from the robot. Additionally, this subsection aims to motivate the problems with current RL approaches and explain the functionality principle of our algorithm $\Pi$-learning.

In RL, the Bellman equation~\cite{bellman1966dynamic} is a fundamental concept that allows the computation of the value function $V(s)$. In this context, the Bellman equation can be expressed as follows:

\begin{equation}\label{C2eq:01}
V(s)=R(s,a,s')+\gamma V(s')
\end{equation}

\noindent Here, $V(s)$ is the value of the state s, $R(s, a, s')$ is the reward for transitioning from $s$ to $s'$ while taking action a, $\mathrm{\gamma}$ is a discount factor and $V(s')$ is the value of the next state. When a policy $\pi$ is given, then the value function can be expressed as follows:

\begin{equation}\label{C2eq:02}
V^{\pi}(s)=R(s,a,s')+\gamma V^{\pi}(s')
\end{equation}

\noindent This form of the Bellman equation works when the environment is deterministic. However, when the environment presents stochastic behaviors, such as the one presented by the bag, it is necessary to add the probability function $P(s'|s,a)$:

\begin{equation}\label{C2eq:03}
V^{\pi}(s)=\sum_{s'}^{}P(s'| s,a)(R(s,a,s')+\gamma V^{\pi}(s'))
\end{equation}

\noindent Despite the equation above includes the stochasticity of the environment, it still needs to consider when the policy is stochastic and not deterministic. Hence, to include a stochastic policy: 

\begin{equation}\label{C2eq:04}
\begin{split}
V^{\pi}(s)=\sum_{a}^{}\pi(a|s)\sum_{s'}^{}P(s'| s,a)(R(s,a,s')\\
+\gamma V^{\pi}(s'))
\end{split}
\end{equation}

\noindent The equation above is known as the Bellman expectation equation, and it can be rewritten in its expectation form: 

\begin{equation}\label{C2eq:05}
V^{\pi}(s) = \mathbb{E}_{\substack{s' \sim P \\ a \sim \pi}} \Bigg[R(s,a,s') + \gamma V^{\pi}(s')\Bigg]
\end{equation}

\noindent In equation~\eqref{C2eq:05}, it can be observed that $V(s)$ evaluates the whole value of the state. However, it does not evaluate each action independently, and this is a limitation that can be solved by calculating the quality $Q(s,a)$ of every action-state pair.  Then, the Bellman equation of the Q function can be expressed as follows:

\begin{equation}\label{C2eq:06}
Q(s)=R(s,a,s')+\gamma Q(s')
\end{equation}

\noindent Here, $Q(s,a)$ is the Q value of an action-state pair, $R(s, a, s')$ is the reward for transitioning from $s$ to $s'$ while taking action $a$, $\mathrm{\gamma}$ is a discount factor and $Q(s',a')$ is the Q value of the next state's action $a'$. When a policy $\pi$ is given, then the Q-function can be expressed as follows:

\begin{equation}\label{C2eq:07}
Q^{\pi}(s,a)=R(s,a,s')+\gamma Q^{\pi}(s',a')
\end{equation}

\noindent This form of the Bellman equation works when the environment is deterministic. However, when the environment presents stochastic behaviors such as the one presented by the bag while being manipulated, it is necessary to add the probability function $P(s'|s,a)$:

\begin{equation}\label{C2eq:08}
\begin{split}
Q^{\pi}(s)=\sum_{s'}^{}P(s'| s,a)(R(s,a,s')\\
+\gamma Q^{\pi}(s',a'))
\end{split}
\end{equation}

\noindent Despite the equation above includes the stochasticity of the environment, it still needs to consider when the policy is stochastic and not deterministic. Hence, to include a stochastic policy: 

\begin{equation}\label{C2eq:09}
\begin{split}
Q^{\pi}(s,a)=\sum_{s'}^{}P(s'| s,a)(R(s,a,s')\\
+\gamma\sum_{a}^{}\pi(a|s)Q^{\pi}(s',a'))
\end{split}
\end{equation}

\noindent The equation above is known as the Bellman expectation equation of the Q-function, and it can be rewritten in its expectation form: 

\begin{equation}\label{C2eq:010}
\begin{split}
Q^{\pi}(s) = \mathbb{E}_{s' \sim P} \Bigg[R(s,a,s')
+ \gamma \mathbb{E}_{a' \sim \pi}Q^{\pi}(s',a')\Bigg]
\end{split}
\end{equation}

\noindent In general, the value and Q function Bellman equations are used for calculating the value of the state and the quality of the Q value's state-action pairs, respectively. These equations are based on the dynamics of the environment given by $P(s'|s,a)$, a policy $\pi$, and the reward function $R(s,a,s')$. The Bellman optimality equation expresses the expected maximum or total reward that can be achieved from a given state based on the value function (Eq.~\eqref{C2eq:04}). The equation defines the relationship between the value of a state and the values of its neighboring states. The Bellman optimality equation of the value function is given by the following:

\begin{equation}\label{C2eq:011}
V^{*}(s)=\max\limits_{a}\sum_{s'}^{}P(s'| s,a)(R(s,a,s')+\gamma V^{*}(s'))
\end{equation}

\noindent The preceding equation states that the optimal value of a state is the maximum expected value obtained by taking the best action in the current state. Moreover, this equation considers the expected values of the resulting states. On the other hand, the Bellman optimality equation of the Q function can also be calculated under the same logic and is denoted as follows:

\begin{equation}\label{C2eq:012}
\begin{split}
Q^{*}(s,a)=\sum_{s'}^{}P(s'| s,a)(R(s,a,s')\\
+\gamma \max\limits_{a'} Q^{*}(s',a'))    
\end{split}
\end{equation}

\noindent Additionally, the relationship between the value and Q functions is that the value function can be derived from the Q function by selecting the maximum Q value for each state. In order to calculate the value function based on the Q function, the maximum Q value over all possible actions in a given state is selected. Consequently, for each state, the action that maximizes the Q value is selected, which becomes the value of that state. Therefore, the value V(s) is equal to the maximum Q value for that a given state s. Then, the relationship can be expressed as follows:

\begin{equation}\label{C2eq:013}
V^{*}(s) = \max\limits_{a} Q^{*}(s, a)
\end{equation}

\noindent Following the logic from the previous sentence, if the maximum value of $V^{*}(s)$ corresponds to the maximum value of $Q^{*}(s,a)$, then the Q function can be derived from the value function by substituting Eq.~\eqref{C2eq:013} in Eq.~\eqref{C2eq:012}, then:

\begin{equation}\label{C2eq:014}
Q^{*}(s,a)=\sum_{s'}^{}P(s'| s,a)(R(s,a,s')+\gamma V^{*}(s))
\end{equation}

In general, the relationship between the value and Q functions is that the value function can be derived from the Q function by selecting the maximum Q value for each state. At the same time, the Q function can be derived from the value function by using the Bellman equation. However, as mentioned before, the model of the bag is unknown. Aiming to solve these sorts of problems, Mote Carlo (MC) methods are algorithms designed to estimate the value function and find the dynamics of the environment through interaction (model-free)~\cite{browne2012survey}. The expected value of the value function of a given state can be approximated by visiting that state for $N$ times and obtaining the total return:

\begin{equation}\label{C2eq:016}
V(s) \approx \frac{1}{N}\sum_{i}^{N}R_i 
\end{equation}

\noindent Here, $R_i$ is the total return. In order to establish a balance between computational efficiency and memory requirements, instead of using the arithmetic mean as in Eq.~\eqref{C2eq:016}, the incremental mean is more commonly used, and it is expressed as follows:

\begin{equation}\label{C2eq:017}
V(s_t) = V(s_t) + \alpha(R_t-V(s_t)),
\end{equation}

\noindent where $\alpha=\frac{1}{N_t}$.  MC methods also focus on solving the problem of estimating the Q function for a given policy. Hence, the following expression can be deduced:

\begin{equation}\label{C2eq:018}
Q(s_t,a_t) = Q(s_t,a_t) + \alpha(R_t-Q(s_t,a_t))
\end{equation}

Despite the potential of MC methods, their working principle design relies on reaching a terminal state to approximate a value function. A drawback is that if the episode is too long, it means costly computation. Consequently, any MC method would not be a proper fit for the bagging task. In this context, Sutton~\cite{sutton1998introduction} proposed an alternative that balances Bellman equation methods and MC methods. The approach is known as Temporal difference TD. These approaches learn by bootstrapping, meaning that instead of waiting till the end of an episode, they updated their value estimation based on the following:

\begin{equation}\label{C2eq:019}
V(s) \approx R(s,a) +\gamma V(s'),
\end{equation}

\noindent where $R(s,a)$ is the immediate reward obtained after performing action $a$ from state $s$. The reward function used in this work is given by the following:

\begin{equation}\label{eq:6}
\begin{split}
    R(s,a)= 
\begin{cases}
    \frac{A_{b_{max}}}{A_{bag}},& s=s_0 \text{ or } s=s_1\\
    \frac{A{o_{max}}}{A_{o}},& s=s_2\\
    1,& s=s_3 \text{, and the object is}\\
    &\text{at the center of the opening}\\
    1,& s=s_4\\
    -0.1& s=s_5\\
    0,&\text{otherwise},
\end{cases}
\end{split}
\end{equation}

\noindent where $A_{b_{max}}$ is the maximum area of the bag when it is unfolded, $A_{bag}$ is the area of the bag after executing an action, $A{o_{max}}$ is the maximum area that the bag's opening can reach, and $A_{o}$ is the bag's opening area after executing an action. 

In a similar manner in which the average mean was replaced for the average mean from Eq.~\eqref{C2eq:017}, then Eq.~\eqref{C2eq:019} can be expressed as follows:

\begin{equation}\label{C2eq:020}
V(s) = V(s) + \alpha(R(s,a)+\gamma V(s')-V(s))
\end{equation}

\noindent The last expression, known as the TD estimation rule, can be used to estimate the value of the state. It is also possible to apply the same logic to the Q function in order to have an optimal policy based on that estimate. The TD estimate rule of the Q function is given by the following:

\begin{equation}\label{C2eq:021}
\begin{split}
Q(s,a) = Q(s,a) + \alpha(R(s,a)\\
+\gamma Q(s',a)-Q(s,a))    
\end{split}
\end{equation}

As previously described, this work proposes the use of several primitive actions, meaning that the action space is discrete. Among several approaches, QL is a popular off-policy RL algorithm that learns an MDP in discrete environments. This algorithm is also model-free because it does not require any previous model of the environment. QL is designed to update the quality values $Q$ of a state-action combination, given by:

\begin{equation}\label{C2eq:1}
Q : A \times S \longrightarrow  \mathbb{R},
\end{equation}

\noindent where $Q$ are the Q-values (usually stored in a table), and they represent the quality of the state-action pair. In other words, the higher the Q-value, the better the action for that state.  Then, a Q-table is necessary to represent each Q-value such that QL updates the state-action pair by using the following:

\begin{equation}\label{C2eq:2}
\begin{split}
Q(s,a) \longleftarrow Q(s,a)+ \alpha[R(s,a)\\
+\gamma \max_{a^{'}}Q(s',a')-Q(s,a)],
\end{split}
\end{equation}

\noindent where $\alpha$ is the learning rate and $\gamma$ is the discount factor. The discount factor is usually a value between 0 and 1 $(0 \leq \gamma \leq 1)$  that balances the importance the agent puts on future rewards rather than immediate rewards. According to Eq.~\eqref{C2eq:2}, the state-action pair is updated based on the next state $s'$ even when that state has not been explored, which is why QL is considered an off-policy method.

Despite the popularity of Q-learning and its successful implementation in multiple fields, when it comes to the problem defined in section~\ref{sec:problem}, there exist several drawbacks. The first one is related to the size of the exploration space, where considering eight primitive actions, 81 pose points, and four possible states in the case of Fig.~\ref{fig:perception}, it would be necessary to explore a total of 2592 primitive action-pose point pairs. This approach lacks practical significance because it would involve a long training time in the real world. 

Additionally, the dependency of Q-learning on the next state value would also involve significant exploration to achieve stable convergence.  This is because of the dynamics of the bag, which despite executing the best action, it would take repeating the same action until the bag's state transitions. More specifically, while for the best action $a$ given the state $s_t$, the bag may transition to $s_{t+1}$, it may also stay in the same state $s_t$ due to the manner the problem was defined (see Eq.~\eqref{eq:a1}) and for the characteristics of the environment, which also include failures in the real world that could contaminate the training. 

Aiming to solve the drawbacks, the authors propose the following. First, the exploration space is reduced by implementing affordances that define what primitive actions are suitable given the current state. The affordances are obtained from a manually defined set of rules $\Psi$. Then, the robot executes actions with probability $\epsilon$, also known as $\epsilon$-greedy policy (the value of $\epsilon$ controls the exploration of the environment), and gets a reward. The process is repeated for $n$ steps till the training is completed. Let: 

\begin{equation}\label{eq:8}
\begin{split}
\Psi=\{\langle s_0,\tau_{grasp}, \mu_{lift} \rangle,\langle s_1,\tau_{scratch} ,\mu_{drag} \rangle,\\
\langle s_2,\tau_{pick},\mu_{place} \rangle,\langle s_3,\tau_{grasp},\mu_{carry} \rangle
\},
\end{split}
\end{equation}

\noindent be the set of rules that contains the tuples in which each state $s$ is related to its valid actions. Then, the primary affordable actions of the states are:

\begin{equation}\label{eq:9}
\Lambda_{s_j,primary}= \Psi^2\times P_{s_j}, 
\end{equation}

\noindent where $s_j \in \Psi^1$ and $j\in(0,|S|)$. The affordable complementary actions are given by:

\begin{equation}\label{eq:10}
\Lambda_{s_j,complementary}= \Psi^3\times D_{s_j},
\end{equation}

\noindent where $s_j \in \Psi^1$. The set of affordable actions pairs is given by:

\begin{equation}\label{eq:11}
A=\Lambda_{s_j,primary}\times\Lambda{s_j,complementary}
\end{equation}

Aiming to avoid confusion with the Q-learning notation algorithm, we utilize $\Pi$. Algorithm~\ref{alg:alg1} has a function that calculates the state-action value $\Pi$, representing the quality value of an action $a$ given a state $s$:

\begin{equation}\label{eq:4}
\begin{split}
\Pi:S\times A \longrightarrow \mathbb{R}
\end{split}
\end{equation}

At the beginning of the learning, all $\Pi$ values are initialized to zero and stored in a $\Pi$-table. 
During the training process, it is updated with the following:

\begin{equation}\label{eq:5}
\begin{split}
\Pi(s,a) \longleftarrow \frac{\Pi(s,a)+R(s,a)}{m},
\end{split}
\end{equation}

\noindent where $m$ is the number of times that action-state pair $\Pi(s,a)$ has been selected by the agent, $m>0$ and $a\in A$. Here, it is not necessary to utilize $\alpha$ as in Eq.~\eqref{C2eq:018} because the compact state representation combined with the $\Pi$-table allows us to know the number of times $m$ that state has been visited. Hence, the bootstrapping technique is not necessary. In contrast to classical Q-learning~\cite{watkins1992}, which would require exploring a total of 2592 primitive action-pose point pairs (considering 8 primitive actions, 81 pose points, and 4 possible states in the case of Fig.~\ref{fig:perception}), $\Pi$-learning is tailored to handle these conditions, particularly in the context of the bagging task. This is because many actions do not result in state changes due to factors such as incomplete unfolding in state 0 or failure to open the bag in state 1. As a result, the robot must repeat the same action until a transition occurs. To this end, the optimal policy is extracted from $\Pi(s,a)$ with: 

\begin{equation}\label{eq:7}
\begin{split}
\pi^*(s)=\argmax_a[\Pi(s,a)]
\end{split}
\end{equation}

Unlike Q-learning, $\Pi$-learning incorporates equation~\eqref{eq:5}, which is independent of the next state. It also reduces the exploration space by defining rules and pairs of primitive actions using equation~\eqref{eq:11}. For instance, in the scenario depicted in Fig.~\ref{fig:perception}, classical Q-learning would require exploring 2592 actions, while $\Pi$-learning would only necessitate exploring 81 actions. This significant reduction in exploration space aims to reduce training time for $\Pi$-learning. Furthermore, the $\Pi$-table remains unaffected by the next state. This is crucial because the bag assumes different shapes after each action and may or may not transition into the next state. This behavior may cause instability for Q-learning because of its dependence on the next state information. Consequently, the perception module detects state transitions and allows $\Pi$-learning to concentrate on maximizing the reward solely based on the current state.

Algorithm~\ref{alg:alg1} takes the number of training steps $n$ and the set of actions $A$ as the input. First, a $\Pi$-table is generated. Then, after $n$ steps of training, the algorithm returns the optimal policy $\pi^*(s)$. The $\Pi$-learning algorithm is specifically designed to enable learning with a reduced number of states while dealing with a wide range of actions. 

\begin{algorithm}[htbp]
\caption{$\Pi$-learning.}\label{alg:alg1}
\begin{algorithmic}[1]
\REQUIRE Training steps $n$, set of actions $A$
\ENSURE Optimal policy $\pi^*(s)$

\STATE Initialize a $\Pi$-table with zeros
\FOR{$n$ steps}
    \STATE With probability $\epsilon$, select a valid action $a$ from $A$
    \STATE Perform $a$ and calculate the reward with Eq.~(\ref{eq:6})
    \STATE Update $\Pi$-table with Eq.~(\ref{eq:5})
\ENDFOR

\STATE Extract the optimal policy from the $\Pi$-table with Equation~(\ref{eq:7})
\end{algorithmic}
\end{algorithm}

\subsection{Robot controller}

The robot controller module coordinates the continuous actions of the robot in a precise manner. To accomplish this, the module inputs any primitive action and translates it into continuous actions that the robot can handle. Since the perception module provides the grasping points of the bag, and the learning module generates a sequence of primitive actions given a state, the robot controller module can make the robot interact with the environment. This information is managed through the Robot Operating System (ROS).

\begin{figure*}[!ht]
\centering
\includegraphics[width=5.2in,scale=1.0]{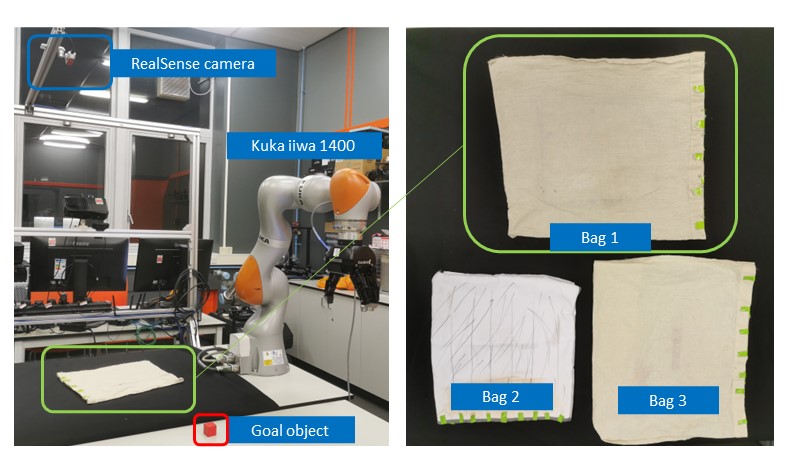}
\caption{The left side of the figure illustrates the experimental setup comprising an object to be bagged (red cube), a Kuka® iiwa 1400™ robot, and an Intel® RealSense™. On the left side are the three bags used during the experiments.  }
\label{fig:setup}
\end{figure*}

\section{Experimental setup}
\label{sec:expsetup}

The experimental setup to validate our framework empirically includes three bags (Table~\ref{table:bag_characteristics}), the object to be bagged (red cube), a Kuka® iiwa 1400™ robot, and an Intel® RealSense™ camera (Fig.~\ref{fig:setup}). The task's goal is to train the robot to learn to bag the object (red cube). The experiments start by training several agents to learn how to handle ``Bag 1'' in the real world for 10, 30, 50, and 100 training steps for each state of the task (unfolding, opening, placing the piece, and carrying), totaling 40, 120, 200 and 400 total training steps, respectively. Then, we compare the performance of our framework against the following state-of-the-art algorithms using the implementations from the stable-baselines~\cite{{stable-baselines}}: dueling deep Q-network (DQN)~\cite{wang2016dueling} and synchronous
actor-critic (A2C)~\cite{mnih2016asynchronous} two robust and well-tested algorithms for discrete action spaces. When the training is finished, we run 10 attempts for each agent and count the number of times the bagging task is successfully finished for each step. Additionally, 10 attempts are performed to calculate the success rate of all the algorithms when starting from the unfolding and opening steps. With this experiment, we aim to discover if our framework can learn better than the baseline algorithms.

\begin{table*}[ht]
    \centering
    \caption{Characteristics and parameters of the bags used in the experiments.}
    \label{table:bag_characteristics}
    \small
    \begin{tabular}{@{}cccccccc@{}}
        \toprule
        \textbf{Name} & \textbf{Opening length} & \textbf{Bag width} & \textbf{Material} & $A_{th}$ & $A_{oth}$ & $A_{b_{max}}$ & $A_{o_{max}}$ \\
        \midrule
        Bag 1 & 30 cm & 35 cm & Cotton & 25000 & 150 & 34000 & 3900\\
        Bag 2 & 25 cm & 25 cm & Polyester & 18000 & 50 & 28000 & 3200\\
        Bag 3 & 33 cm & 26 cm & Cotton & 25000 & 150 & 34000 & 3900\\
        \bottomrule
    \end{tabular}
\end{table*}
The trained agent with the highest success rate is used to test the generalization capacities of the framework. To evaluate the framework's proficiency in handling the task from a different starting position, we change the position and orientation of ``Bag 1'' twice. Then, the framework is tested on ``Bag 2'' and ``Bag 3'' for 10 attempts. 

\begin{figure*}[!ht]
    \centering
    \begin{minipage}{0.45\textwidth}
        \subfloat[]{\includegraphics[width=\linewidth]{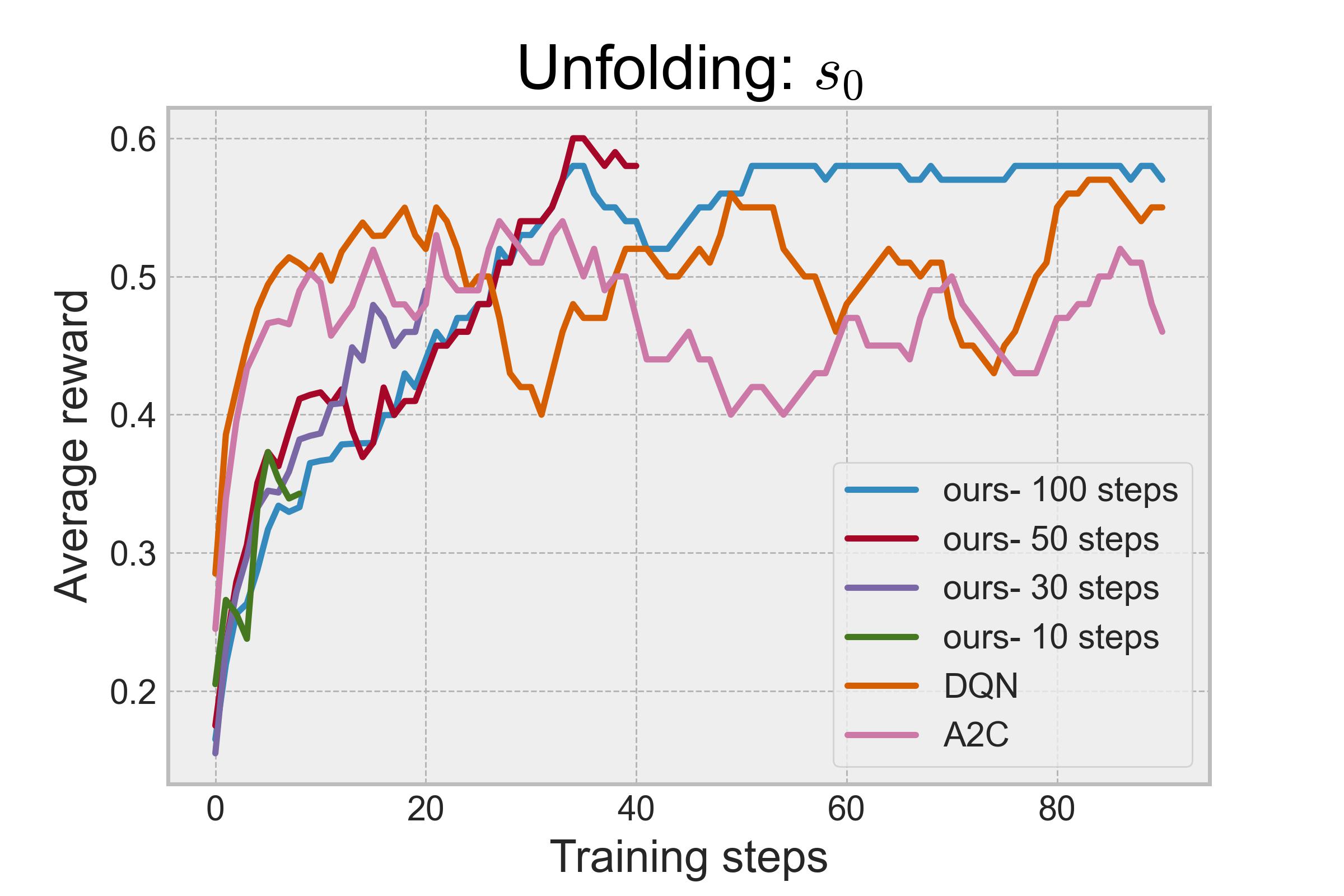}%
        \label{fig:results_a}}
    \end{minipage}
    \hspace{0.05\textwidth}
    \begin{minipage}{0.45\textwidth}
        \subfloat[]{\includegraphics[width=\linewidth]{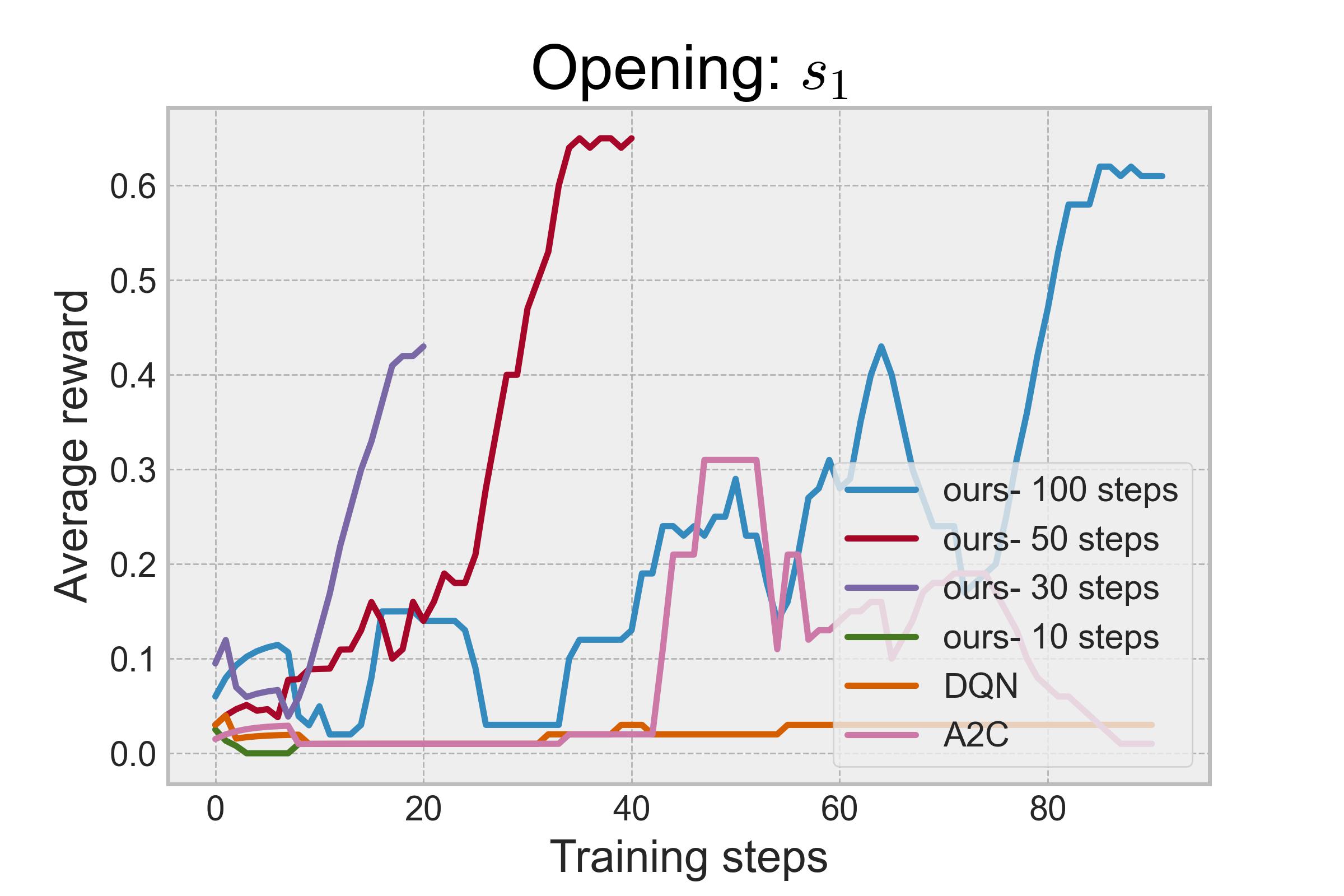}%
        \label{fig:results_b}}
    \end{minipage}
    \vspace{0.05\textwidth}
    \begin{minipage}{0.45\textwidth}
        \subfloat[]{\includegraphics[width=\linewidth]{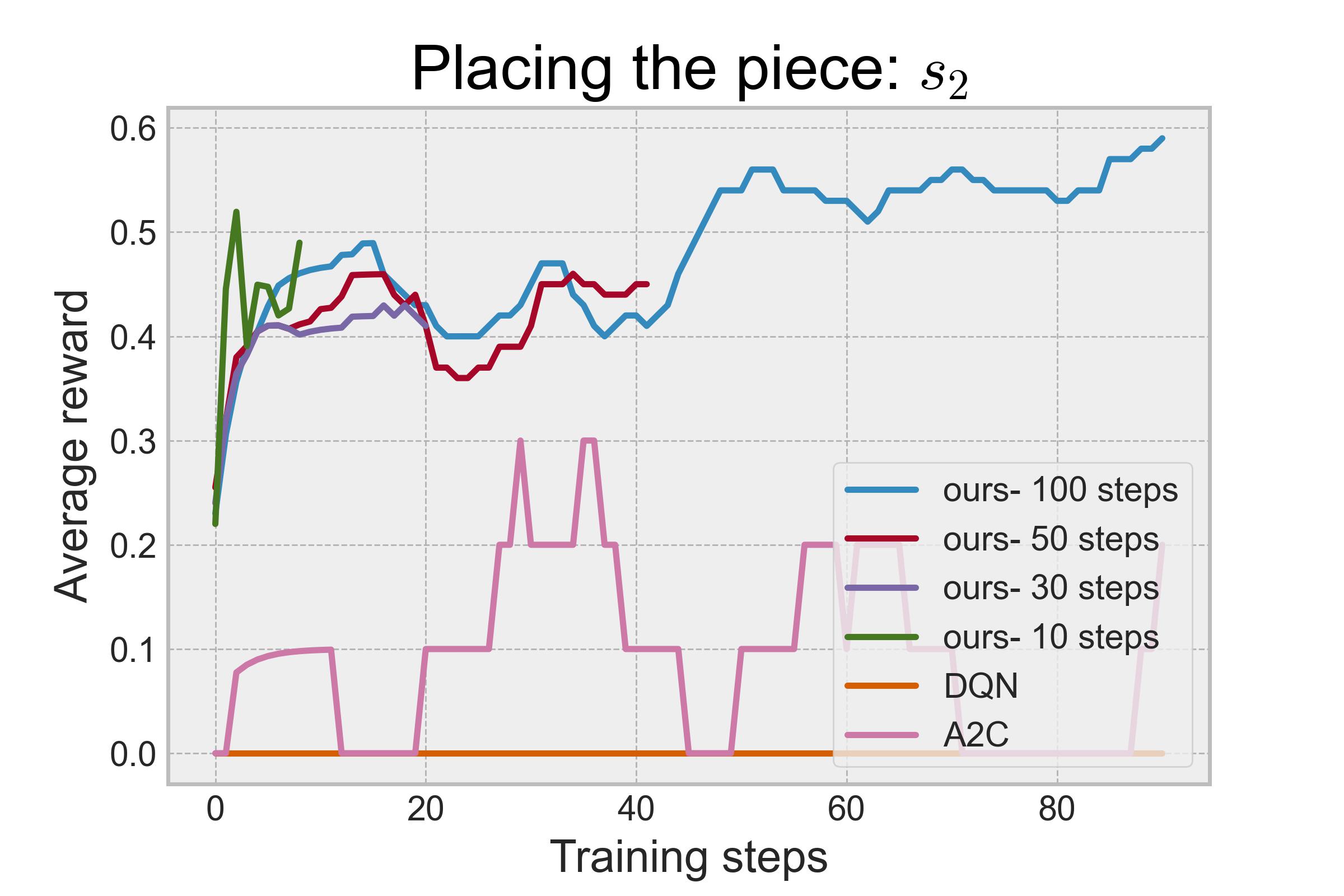}%
        \label{fig:results_c}}
    \end{minipage}
    \hspace{0.05\textwidth}
    \begin{minipage}{0.45\textwidth}
        \subfloat[]{\includegraphics[width=\linewidth]{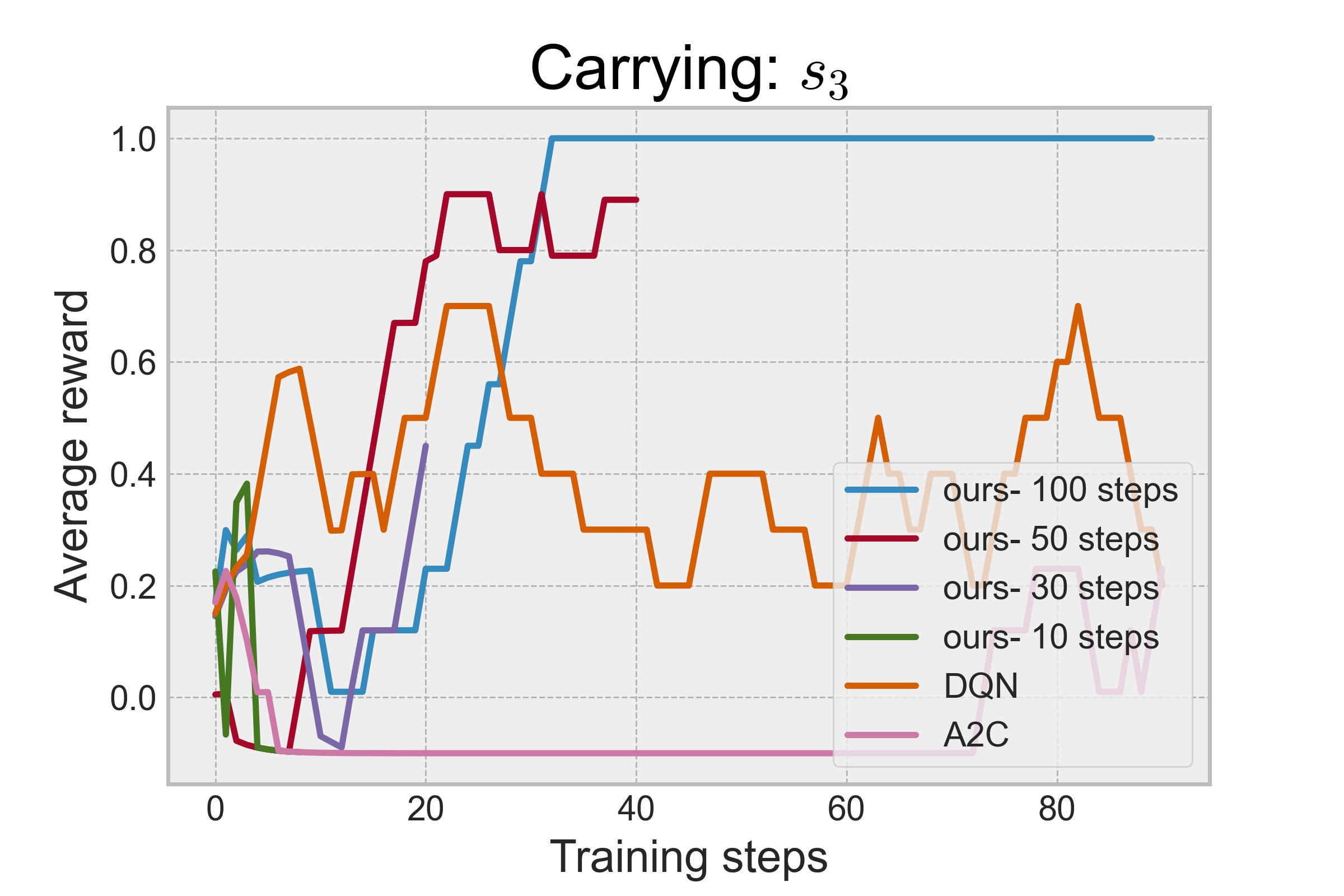}%
        \label{fig:results_d}}
    \end{minipage}
    \caption{The learning curves above display the results of our experiments. (a) shows the learning curve progress of the unfolding step, demonstrating that our approach converges after 100 training steps while DQN and A2C struggle to do so. In (c), our approach was the only one to converge after training for 100 and 50 training steps. In (d), A2C and DQN failed to find a solution while our approach converged. Lastly, in (e), our approach trained for 100 steps and converged to the highest value.}
    \label{fig:results}
\end{figure*}

\section{Results}
\label{sec:results}

\begin{figure*}[!t]
\centering
\smallskip
\begin{tabular}{ c }
        \subfloat[]{\includegraphics[width=1.0\textwidth]{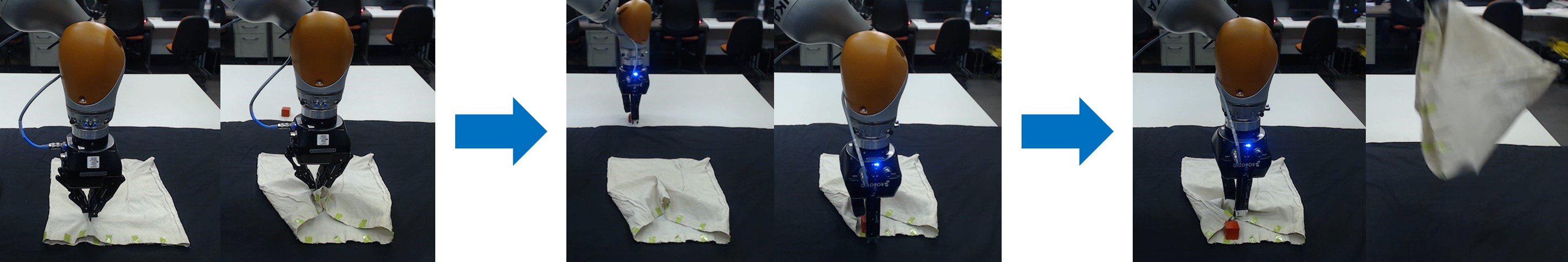}%
        \label{fig:exp1}}\\
        \subfloat[]{\includegraphics[width=1.0\textwidth]{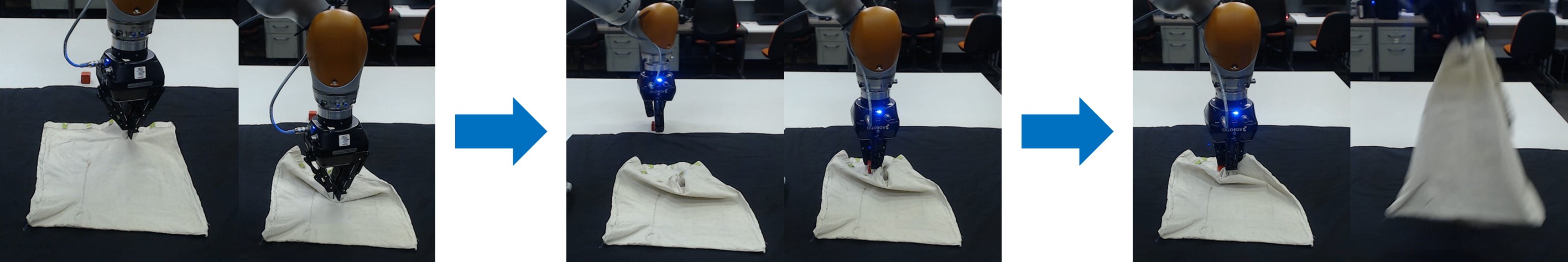}%
        \label{fig:exp2}}\\
        \end{tabular}    
\caption{The robot performing the bagging task with two different bags. In (a), the bag's opening faces the camera's view. In (b), the bag's opening is facing the opposite direction of the camera's view. The robot successfully completed the tasks in both cases with different orientations of the bag.}
\label{fig:exps}
\end{figure*}

\begin{figure*}[!t]
\centering
\smallskip
\begin{tabular}{ c }
        \subfloat[]{\includegraphics[width=0.95\textwidth]{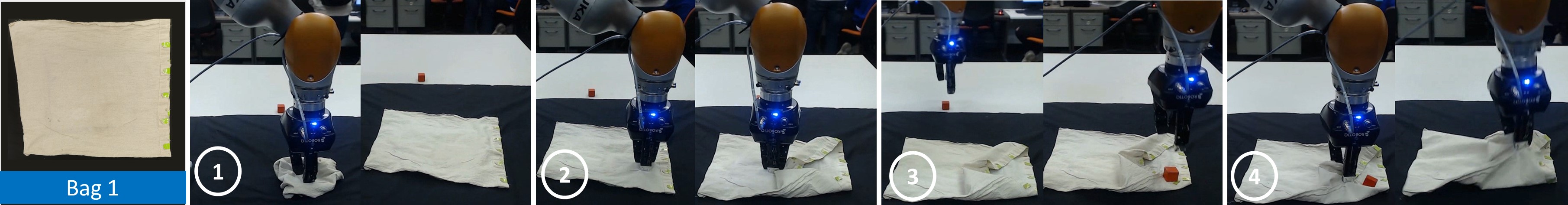}%
        \label{fig:exp3}}\\
        \subfloat[]{\includegraphics[width=0.95\textwidth]{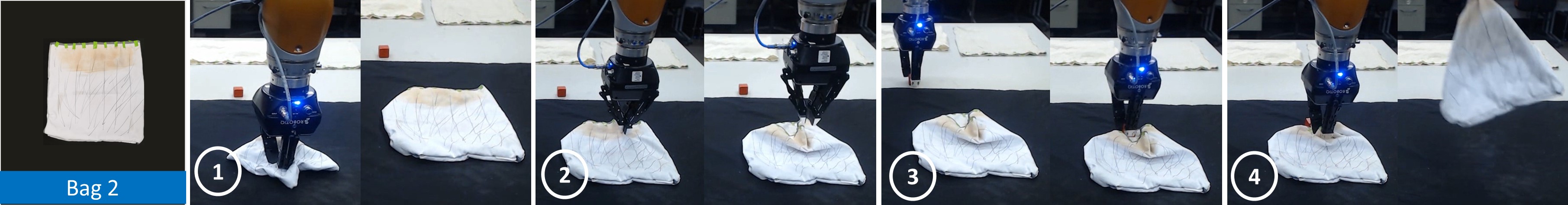}%
        \label{fig:exp4}}\\
        \subfloat[]{\includegraphics[width=0.95\textwidth]{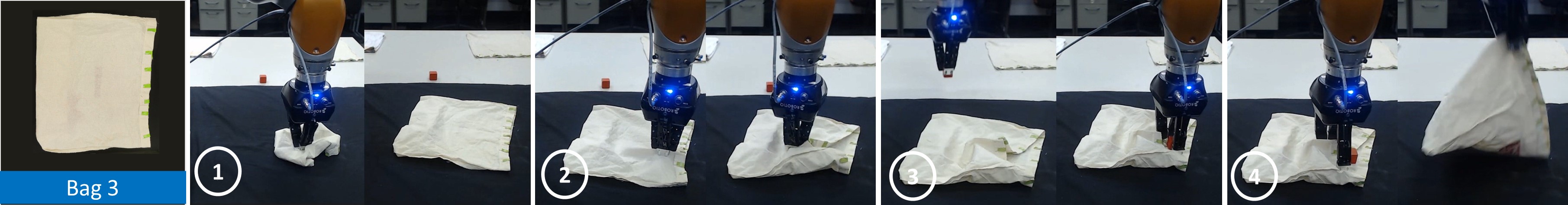}%
        \label{fig:exp5}}\\
        \end{tabular}    
\caption{The robot performing the bagging task starting from step 1 with three different-sized bags. In (a), the robot performs the bagging task with ``Bag 1", used for training with our framework. In (b), the robot using the framework after training performs the bagging task with ``Bag 2", a smaller bag made of polyester. In (c), the robot using the framework after training performs the bagging task with ``Bag 3", a bag made of cotton and also with a different size of ``Bag 1".}
\label{fig:different_bags}
\end{figure*}

This section presents the results of the experiments previously described. Firstly, the learning progress of the framework and the state-of-the-art algorithms for each step of the task (unfolding, opening, placing the piece, and carrying) is illustrated in Fig.~\ref{fig:results}. Secondly, Table~\ref{table:results_training} shows the total reward obtained by each algorithm, followed by Table~\ref{table:results_training_perstep}, which contains the reward obtained by all the approaches for each step of the bagging task. Table~\ref{table:success_perstep_bag_1} shows the success rates from performing the bagging task 10 times with ``Bag 1" for each step of the task, as well as the success rates starting from step 1 (opening) and from step 2 (unfolding). Then, Fig.~\ref{fig:exps} shows the robot performing the task in different initial positions with ``Bag 1". Fig.~\ref{fig:different_bags} illustrates the robot performing the bagging task with ``Bag 2" and `Bag 3". Lastly, the success rates of handling all the bags are summarized in Table~\ref{table:success}.

The learning curves in Fig.~\ref{fig:results_a} show the progress of the agent learning to unfold the bag, which consists in selecting a grasping point, lifting the bag, and dropping it till it is unfolded. Our framework running for 100 training steps converged in around 50 training steps, while DQN and A2C demonstrated an unstable learning behavior by struggling to converge. The framework running 10 and 30 training steps shows that the agent requires more exploration. On the other hand, our framework learning for 50 training steps indicates that 50 is the minimum number of training steps required to find the best grasping and lifting positions for our approach.

The learning curves in Fig.~\ref{fig:results_b} illustrate the learning progress of the agents for the opening step, which involves grasping one layer of the bag and dragging it to another point of the bag. The reward during this step is equal to the total area of the opening. The framework that ran for 100 training steps converged in approximately 80 training steps, while the one that ran for 50 training steps found the best solution in around 40 training steps. This is due to the stochastic nature of the exploration, which randomly found a better action in an early stage of the learning process for the 50 training steps experiment. DQN and A2C fell into a local minimum, and their learning could not progress. Our framework running for 10 and 30 training steps could not explore the environment enough to find an optimal policy.

The learning curves in Fig.~\ref{fig:results_c} show the progress of the agent learning to place the goal object (red cube) in the bag’s opening such that the closer the robot places the goal object, the higher the reward. Our framework running for 10, 30, 50, and 100 training steps converged faster than the previous steps because this step is more straightforward than the previous ones by only including nine placing points in the bag’s opening. The framework running for 10, 30, 50, and 100 could converge, while DQN and A2C could not find a solution.  

\begin{table*}[h!]
    \centering
    \caption{Total reward obtained by our framework and stable-baselines after training.}
    \label{table:results_training}
    \begin{tabular}{cccccc}
        \toprule
        \textbf{Approach} & \textbf{Total reward} & \textbf{Training time} & \textbf{Total training steps} \\
        \midrule
        Ours (100) & 1.9608 & 173 min. & 400 (100 for each step of the task) \\
        Ours (50)  & 1.7    & 95 min.  & 200 (50 for each step of the task)  \\
        Ours (30)  & 0.99   & 62 min.  & 120 (30 for each step of the task)  \\
        Ours (10)  & 0.2475 & 18 min.  & 40 (10 for each step of the task)   \\
        DQN        & 1.03   & 198 min. & 400 (100 for each step of the task) \\
        A2C        & 0.6488 & 186 min. & 400 (100 for each step of the task) \\
        \bottomrule
    \end{tabular}
\end{table*}

The learning curves in Fig.~\ref{fig:results_d} show the learning progress of the agents for the carrying task. In this step, if the robot executes the carrying action and the red cube is not on the table after that, the reward is equal to 1 and -0.1 otherwise. The framework that runs for 50 and 100 training steps could converge to a stable plateau while running for 10 and 30 could not result in the right grasping point for carrying the bag. DQN fell into a local minimum and could not find a solution. A2C could find a better grasping point to carry the bag. However, the learning curve shows that A2C struggles to converge. 

\begin{table*}[ht]
    \centering
    \caption{Reward obtained by our framework and stable-baselines after training per step.}
    \label{table:results_training_perstep}
    \small
    \begin{tabular}{ccccc}
        \toprule
        \textbf{Approach} & \textbf{Reward step 1} & \textbf{Reward step 2} & \textbf{Reward step 3} & \textbf{Reward step 4} \\
        \midrule
        Ours (100) & 0.5   & 0.241 & 0.49  & 0.73  \\
        Ours (50)  & 0.46  & 0.28  & 0.44  & 0.52  \\
        Ours (30)  & 0.39  & 0.248 & 0.43  & 0.22  \\
        Ours (10)  & 0.366 & 0.014 & 0.5   & 0.11  \\
        DQN        & 0.51  & 0.02  & 0.12  & 0.38  \\
        A2C        & 0.47  & 0.07  & 0.1   & 0.009 \\
        \bottomrule
    \end{tabular}
\end{table*}

The success rates of all the approaches are summarized in Table~\ref{table:success}, in which 10 attempts were performed for each step of the bagging task. For step 1, Ours(100) and Ours(50) reached the highest success rate with 70\%, followed by DQN with 40\%. The lowest success rates were achieved by Ours(30) and A2C with 0\% and 20\%, respectively. For step 2, A2C, DQN, and Ours(10) presented the lowest success rates, which demonstrates that step 2 is the most difficult to learn. For step 3, all the approaches reached a success rate equal to or superior to 90\%, which demonstrates that this step is the easiest to learn. In the last step, Ours(10) had the lowest performance with 60\% while the rest of the approaches reached a success rate equal to or superior to 90\%. Additionally, 10 attempts were carried out from step 1 (unfolding) and step 2 (opening), which for Ours(100) resulted in 60\% and 80\% of success rates, respectively. It can be observed that the low success rates of Ours(10), DQN, and A2C are because of getting stuck on step 2. Moreover, the difficulty increases when the task is started from step 1, which reflects in the performance of all the agents. 

\begin{table*}[ht]
    \centering
    \caption{Success rate of the framework and stable-baselines after trainin.}
    \label{table:success_perstep_bag_1}
    \small 
    \begin{tabular}{@{}lcccccc@{}}
        \toprule
        \textbf{Experiment} & \textbf{Step 1} & \textbf{Step 2} & \textbf{Step 3} & \textbf{Step 4} & \textbf{Success rate} & \textbf{Success rate} \\
        \textbf{``bag 1"}& & & & & \textbf{from Step 1} & \textbf{from Step 2} \\
        \midrule
        Ours (100) & 7/10 & 9/10 & 9/10 & 10/10 & 6/10 & 8/10 \\
        Ours (50)  & 7/10 & 7/10 & 9/10 & 10/10 & 5/10 & 6/10 \\
        Ours (30)  & 2/10 & 4/10 & 10/10 & 10/10 & 1/10 & 4/10 \\
        Ours (10)  & 0/10 & 0/10 & 10/10 & 6/10  & 0/10 & 0/10 \\
        DQN        & 4/10 & 1/10 & 10/10 & 9/10  & 0/10 & 2/10 \\
        A2C        & 2/10 & 1/10 & 10/10 & 10/10 & 0/10 & 1/10 \\
        \bottomrule
    \end{tabular}
\end{table*}

The total average reward obtained by all the approaches is summarized in Table~\ref{table:results_training}, where the three approaches that collected the highest rewards are Ours(100), Ours(50), and DQN with 1.9608, 1.7 and 1.03, respectively. When it comes to the total average reward collected for each step, Table~\ref{table:results_training_perstep} shows that for step 1, DQN collected the highest reward of 0.51, followed by Ours(100), which collected 0.5. For step 2, Ours(50) collected the highest reward of 0.28, followed by Ours(100) with 0.241. In step 3, the highest reward was obtained by Ours(100). For step 4, the highest reward was obtained by Ours(100), followed by Ours(50) and DQN. In general, the stable-baselines DQN and A2C presented problems learning step 2 (opening), while the rewards collected for all the approaches in step 1 are almost the same. 

\begin{table*}[ht]
    \centering
    \caption{Success rate of the framework and stable-baselines after training per step.}
    \label{table:success}
    \small
    \begin{tabular}{@{}lcccccc@{}}
        \toprule
        \textbf{Experiment} & \textbf{Step 1} & \textbf{Step 2} & \textbf{Step 3} & \textbf{Step 4} & \textbf{Success rate} & \textbf{Success rate} \\
         & & & & & \textbf{from Step 1} & \textbf{from Step 2} \\
        \midrule
        Bag 1 & 6/10 & 8/10 & 9/10 & 9/10 & 6/10 & 8/10 \\
        Bag 2 & 4/10 & 6/10 & 10/10 & 8/10 & 2/10 & 5/10 \\
        Bag 3 & 5/10 & 8/10 & 10/10 & 9/10 & 3/10 & 7/10 \\
        \bottomrule
    \end{tabular}
\end{table*}

The last experiment tested the generalization capabilities of the framework. The success rates are summarized in Table~\ref{table:success_perstep_bag_1}. Fig.~\ref{fig:exps} illustrates the robot performing the bagging task with ``Bag 1'' starting from two different positions and orientations. Despite the change in the initial position, the framework was capable of finishing the task. This is because our approach focuses on the state and grasping points with respect to the bag, which is independent of the pose of the bag in the global workspace frame. Fig.~\ref{fig:exp3} shows the robot performing the bagging task with ``Bag 2'', which had a success rate of 20\% when starting from step 1 and 50\% when starting from step 2. The robot performing the 
bagging task with ``Bag 3'' had a success rate of 30\%  and 70\% when starting from step 1 and step 2, respectively. Most of the failures are because of the robot being unable to open the bag and not being able to unfold it.

\section{Discussion}
\label{sec:discussion}

Prior to the current work, we implemented the DQN and A2C algorithms aiming to solve the problem of learning bagging in the real world. However, we observed low performance during the bagging task with DQN and A2C (refer to Table~\ref{table:results_training}), which we attribute to the limited number of training steps (400 training steps in total). Achieving better results with these algorithms would likely require implementing them in a simulated environment. However, this approach presents challenges, such as the reality-to-simulation gap discussed in Section~\ref{sec:rel} and the need to accurately simulate the physical properties of the bag. Furthermore, DQN and A2C algorithms typically require thousands to millions of steps to achieve stable learning. For instance, Hester et al.~\cite{hester2018deep} demonstrated that the Deep Q-Network from Demonstrations (DQNfD) required 1 million steps to achieve satisfactory scores in their experiments. DQN took 84 to 85 million steps for similar performance in the same application. This problem led to the design of the $\Pi$-learning algorithm.

Additionally, after completing the learning phase with our framework, the robot performed 100 attempts using each approach for the bagging task with ``Bag 1" (refer to Fig.~\ref{fig:exp3}). The success rates are summarized in Table~\ref{table:results_training}, in which it can be appreciated that the main reason for the failures was the robot's inability to complete the unfolding (step 1) or opening of the bag (step 2). This highlights the need for further improvements, such as a more robust unfolding strategy and enhanced measurement accuracy of the camera. Improved accuracy of the camera's measurements would allow the robot's gripper to reach the surface of the bag's layer more precisely, as even a difference of 1 mm caused the robot to grasp both layers or fail to grasp any layer at all.

The decision to not use continuous RL algorithms, such as Deep Deterministic Policy Gradient (DDPG)~\cite{lillicrap2015continuous}, or Soft Actor-Critic (SAC)~\cite{haarnoja2018soft}, is motivated by the aim of preventing dangerous behaviors of the robot in our experimental setup, particularly during the learning stage. Unlike continuous RL algorithms, which may lead to collisions and undesired behaviors due to their inherent exploration process, defining primitive actions and employing a discrete action-selection approach helped to prevent such incidents specifically for the task proposed in this paper.

\section{Conclusion}
\label{sec:conclusions}

This paper presented an efficient learning framework for a robot manipulator to acquire the bagging task. Our real-world learning robot-bagging framework has been empirically validated. Leveraging our novel reinforcement learning algorithm $\Pi$-learning, this framework enables efficient learning of the bagging task in real-world scenarios. After training for a total of 400 steps, which took approximately three hours, the framework achieved a success rate of 60\% and 80\% when starting from the unfolding or opening step, respectively. Additionally, the framework demonstrated generalization capabilities across different bags.

However, there are certain limitations to the proposed framework. For instance, the framework's applicability is restricted to bags made from specific materials, such as textile-based bags, and may not be suitable for handling plastic bags. Furthermore, the framework is designed to handle only one object that is smaller than the bag's opening. Additionally, there is room for improvement in the primitive actions of unfolding the bag ($\tau_{grasp}$) and grasping only one layer of the bag ($\tau_{scratch}$), as these actions were the primary causes of the robot's failure in the bagging tasks. Despite these limitations, our framework exhibits the potential to tackle challenging problems such as bag manipulation.

For future work, we have plans to enhance the unfolding and opening routines by incorporating bi-manual robotic manipulation techniques. This advancement would enable the bagging of multiple objects, expanding the framework's capabilities beyond a single object. Additionally, we aim to explore the integration of supervised learning methods, which would facilitate the generalization of the perception module to a wider range of bag types. This extension would enhance the framework's versatility and applicability.  

\bmhead{Acknowledgments}
This work was partially supported by Consejo Nacional de Humanidades, Ciencias y Tecnologías and the Engineering and Physical Sciences Research Council (grant No. EP/X018962/1).







\bibliography{sn-bibliography.bib}

\end{document}